# Bayesian Learning of Loglinear Models for Neural Connectivity

**Kathryn Blackmond Laskey**
Department of Systems Engineering and $C^3I$ Center
George Mason University
Fairfax, VA  22030
klaskey@gmu.edu

**Laura Martignon**
Max Planck Institut für Psychologische Forschung
Center for Adaptive Behavior and Cognition
80802 München, Germany
laura@mpipf-muenchen.mpg.de

## Abstract

This paper presents a Bayesian approach to learning the connectivity structure of a group of neurons from data on configuration frequencies. A major objective of the research is to provide statistical tools for detecting changes in firing patterns with changing stimuli. Our framework is not restricted to the well-understood case of pair interactions, but generalizes the Boltzmann machine model to allow for higher order interactions. The paper applies a Markov Chain Monte Carlo Model Composition ($MC^3$) algorithm to search over connectivity structures and uses Laplace's method to approximate posterior probabilities of structures. Performance of the methods was tested on synthetic data. The models were also applied to data obtained by Vaadia on multi-unit recordings of several neurons in the visual cortex of a rhesus monkey in two different attentional states. Results confirmed the experimenters' conjecture that different attentional states were associated with different interaction structures.

## 1 INTRODUCTION

Simultaneous activation of two or more neurons is, as Aertsen and Grün say, the *unitary event* for information processing in the brain [1]. Since the time of Hebb's [2] fundamental theory it has been understood that complex information processing in the brain arises from the collective interaction of groups of neurons. Experimental advances in the last decade enable the direct study of firing patterns among the spiking events of groups of neurons. Abeles and his coworkers in Jerusalem have developed and applied new measurement methods, and have reported coincidences occurring at fixed time intervals with a much higher probability than chance ([3-6]). Interactions of order higher than two indicate that three or more neurons share a common input. Detecting such higher-order interactions is of great concern to neuroscientists. Most connectionist models assume pairwise interactions among nodes [7-9]. Martignon, et al. [10] developed a family of models capable of representing higher order interactions. Their frequentist approach is subject to well-known difficulties, especially those associated with multiple simultaneous hypothesis tests. The methods reported here were developed to address these problems.

Most previous work on neural networks (e.g., [9]) has taken the connectivity structure as given and focused on learning interaction strength parameters. Recently statistical approaches have become available for structure learning ([11-14]; see also [15]). This work is now being applied to neural networks [16-18].

This paper presents a family of loglinear models capable of capturing interactions of all orders. An algorithm is presented to learn both structure and parameters in a unified Bayesian framework. Each model structure specifies a set of clusters of nodes, and structure-specific parameters represent the directions and strengths of interactions among them. The Bayesian learning algorithm gives high posterior probability to models that are consistent with the data. An advantage of the Bayesian approach is the ease of interpretation of the results of analysis. Results include a probability, given the observations, that an interaction among a set of nodes occurs, and a posterior probability distribution for the strength of the interaction, conditional on its occurrence. Another advantage of the Bayesian approach is the natural way in which uncertainty about structure can be represented and accounted for in data analysis. Studies indicate that explicitly incorporating structural uncertainty improves the ability of models to predict future observations [12]; [19-20].

## 2 THE FAMILY OF MODELS FOR INTERACTIONS

Let $\Lambda$ be a set of nodes labeled 1 through $k$. At a given time each node may be in state 1 (active) or 0 (inactive). The state of node $i$ at a given time is represented by a random variable $X_i$. The lowercase letter $x_i$ is used to denote the state of $X_i$. The state of the set $\Lambda$ is denoted by the random vector $X = (X_1, ..., X_k)$, which can be in one of $2^k$ configurations [1]. The actual configuration at a given time is denoted by $x = (x_1, ..., x_k)$.

---

[1] In this paper we follow the convention that uppercase letters represent random variables, lowercase letters represent specific states of random

A cluster of nodes exhibits a positive (or excitatory) interaction when all nodes in the cluster tend to be active simultaneously and a negative (inhibitory) interaction when simultaneous activity tends not to occur. Our objective is to introduce a mathematical model that allows us to define excitatory and inhibitory interactions in rigorous terms, to identify clusters exhibiting interactions, and to estimate the strength of interaction among nodes in a given cluster.

Let the probability of occurrence of configuration $x$ in a given context be denoted by $p(x)$. The context refers to the situation in which the activity occurs; e.g., the task the animal is performing, the point in the task at which the measurement is taken, etc. If nodes act independently, the probability distribution $p(x)$ is equal to the product of marginal probabilities for the individual neurons:

$$p(x) = p(x_1)\cdots p(x_k). \quad (1)$$

A system of neurons is said to exhibit *interaction* if their joint distribution cannot be characterized by (1). Interactions among neurons can be modeled mathematically by introducing interaction terms into the expression for $p(x)$ (see (2) below). A system of neurons is said to exhibit degree $r$ interaction if at least one interaction term involves $r$ neurons and no interaction term involves more than $r$ neurons. In a system which exhibits degree $r$ interaction, a set of $r$ neurons exhibits *excitatory* (*inhibitory*) interaction if their probability of simultaneous firing is greater (less) than that predicted by interactions of lower order.[2]

To formalize these ideas, let $\Xi$ be a set of clusters of nodes, where each cluster $\xi$ is a subset of $\Lambda$; the empty set $\emptyset$ is required to belong to $\Xi$. Define the random variable $T_\xi$ to have the value 1 if all nodes in cluster $\xi$ are active and 0 if any node in cluster $\xi$ is inactive. That is, $T_\xi$ is the product of the $X_i$ for all $i \in \xi$. By definition, $T_\emptyset$ is equal to 1. The distribution $p(x)$ is assumed to be strictly positive, i.e., all configurations are possible. Thus, its logarithm is well-defined and can be expanded in terms of the components $x_i$ for $i=1,...N$. Let $h(x)$ be the natural logarithm of the probability $p(x)$ of configuration $x$. We say that the nodes in set $\Lambda$ exhibit *interaction structure* $\Xi$ with *interaction parameters* $\{\theta_\xi\}_{\xi \in \Xi}$ if $h(x)$ can be written:

$$h(x) = \log p(x) = \sum_{\xi \in \Xi} \theta_\xi t_\xi . \quad (2)$$

A cluster $\xi$ is said to exhibit excitatory interaction if $\theta_\xi > 0$ and inhibitory interaction if $\theta_\xi < 0$. From (2) it follows that when an excitatory interaction $\theta_\xi > 0$ occurs, the probability of simultaneous firing of the neurons in $\xi$ is greater than what would be the case if $\xi \notin \Xi$ and all other $\theta_\xi$ for $\xi \neq \emptyset$ remained the same.

Any strictly positive probability distribution can be written in the form (2) for $\Xi = 2^\Lambda$. Structures $\Xi$ in which some clusters do not appear are equivalent to the structure $2^\Lambda$ with $\theta_\xi = 0$ for $\xi \notin \Xi$. Martignon et al. [10], [21] studied some special cases of $\Xi$, namely the case of $\Xi = 2^\Lambda$ and the case in which $\Xi$ contains all subsets with cardinality less than a fixed value.[3]

The model (2) is an example of a *loglinear model*, that is, a model in which the logarithm of the probability of an observation is a linear function of some set of statistics (in this case, the $T_\xi$) computed from the observations. It is common in the literature on loglinear models to assume a *hierarchy* constraint [22]. In models of the form (2) this would mean that if $T_\xi \neq 0$ for some cluster $\xi$, then $T_{\xi'}$ would be required to be nonzero for each $\xi' \subset \xi$. The neurobiological context for which our models are being developed precludes this assumption. Neurobiologists are specifically concerned with identifying situations in which simultaneity of activation exists for a cluster $\xi$ of neurons, although its strict subsets are not necessarily simultaneously active. The neurobiological context also precludes restricting the models (2) to those that are *decomposable* ([23-25]). Thus, closed form solutions are not available for either structure or parameter learning.

Let $\theta$ denote the vector of interaction parameters for a structure $\Xi$.[4] To emphasize dependence of the configuration probability on $\theta$ and $\Xi$, we adopt the notation $p(x|\theta,\Xi) = \exp\{h(x|\theta,\Xi)\}$ for the configuration probabilities and their natural logarithms.

This paper is concerned with two issues: identifying the interaction structure $\Xi$ and estimating the interaction strength vector $\theta$. Our approach is Bayesian. Information about $p(x)$ prior to observing any data is represented by a joint probability distribution called the *prior distribution* over $\Xi$ and the $\theta$'s. Observations are used to update this probability distribution to obtain a *posterior distribution* over structures and parameters. The posterior probability of a cluster $\xi$ can be interpreted as the probability that the $r$ nodes in cluster $\xi$ exhibit a degree-$r$ interaction. The posterior distribution for $\theta_\xi$ represents structure-specific information about the magnitude of the interaction. The mean or mode of the posterior distribution can be used as a point estimate of the interaction strength; the standard deviation of the posterior distribution reflects remaining uncertainty about the interaction strength.

## 3. ESTIMATING PARAMETERS FOR AN INTERACTION STRUCTURE

In this section, we consider the problem of using observations to estimate interaction strengths $\{\theta_\xi\}_{\xi \in \Xi}$

---

variables, boldface letters represent vectors, and ordinary letters represent scalars.

[2] This heuristic definition must be modified (see discussion below) when the system also exhibits interactions of order greater than $r$.

[3] In [10] the model was expressed as the *negative* natural logarithm of the configuration probability. Thus, the resulting parameters are the negatives of $\theta$'s obtained here. The decomposition used here has the advantage that positive $\theta$'s represent excitatory interactions and negative $\theta$'s represent inhibitory interactions.

[4] The vector $\theta$ contains $\theta_\xi$ for $\xi \in \Xi$. For notational simplicity the dependence of $\theta$ on the structure $\Xi$ is suppressed.

conditional on a fixed interaction structure $\Xi$. Initial information about $\theta_\xi$ is expressed as a prior probability distribution $g(\theta|\Xi)$. Let $x_1, \ldots, x_N$ be a sample of $N$ independent observations from the distribution $p(x|\theta,\Xi)$. The joint probability of the observed set of configurations, $p(x_1|\theta,\Xi)\cdots p(x_N|\theta,\Xi)$, viewed as a function of $\theta$, is called the *likelihood function* for $\theta$. From (2), it follows that the likelihood function can be written

$$\begin{aligned} L(\theta) &= p(x_1|\theta,\Xi)\cdots p(x_N|\theta,\Xi) \\ &= \exp\left\{\sum_i h(x_i|\theta)\right\} \\ &= \exp\left\{N\sum_{\xi\in\Xi}\theta_\xi \bar{t}_\xi\right\}, \end{aligned} \quad (3)$$

where $\bar{t}_\xi = \frac{1}{N}\sum t_{\xi j}$ is the frequency of simultaneous activation of the nodes in cluster $\xi$. The expected value of $\bar{T}_\xi$ is the probability that $T_\xi=1$, and is often referred to as the marginal function for cluster $\xi$. The observed value $\bar{t}_\xi$ is referred to as the marginal frequency for cluster $\xi$.

The posterior distribution for $\theta$ given the sample $x_1, \ldots, x_N$ is given by:

$$\begin{aligned} &g(\theta|\Xi,x_1,\ldots,x_N) \\ &= Kp(x_1|\theta,\Xi)\cdots p(x_N|\theta,\Xi)g(\theta|\Xi) \end{aligned} \quad (4)$$

where the constant of proportionality $K$ is chosen so that (4) integrates to 1:

$$K = \left(\int_{\theta'} p(x_1|\theta,\Xi)\cdots p(x_N|\theta,\Xi)g(\theta|\Xi)d\theta'\right)^{-1} \quad (5)$$

In this expression, $\theta'$ is the vector of $\theta_\xi$ for nonempty $\xi\in\Xi$. Integration is performed only over parameters that may vary independently. Because the probabilities are constrained to sum to 1, $\theta_\varnothing$ is a function of the other $\theta_\xi$:

$$\theta_\varnothing = -\log\left(\sum_x \exp\left\{\sum_{\substack{\xi\in\Xi\\\xi\neq\varnothing}}\theta_\xi t_\xi(x)\right\}\right) \quad (6)$$

The random vector $\bar{T}$ of marginal frequencies is called a *sufficient statistic* for $\theta$ because the posterior distribution of $\theta$ depends on the data only through $\bar{T}$[5]. Thus, for the purpose of computing the posterior distribution for a fixed structure, no information is lost if the data are summarized by the vector of observed cluster activation frequencies for those clusters belonging to $\Xi$.

A structure $\Xi$ indicates which $\theta_\xi$ are nonzero. The prior distribution $g(\theta|\Xi)$ expresses prior information about the nonzero $\theta_\xi$. We assume that the nonzero $\theta_\xi$ are normally distributed with zero mean and standard deviation $\sigma=2$. That is, we are assuming that the $\theta_\xi$ are symmetrically distributed about zero (i.e., excitatory and inhibitory interactions are equally likely) and that most $\theta_\xi$ lie between -4 and 4. The standard deviation $\sigma=2$ is based on previous experience applying this class of models to other data [10].

The posterior distribution $g(\theta|\Xi,x_1,\ldots,x_N)$ cannot be obtained in closed form. Thus approximation is necessary. Options for approximating the posterior distribution include analytical or sampling-based methods. Because we must estimate posterior distributions for a large number of structures, Monte Carlo methods are infeasibly slow. We therefore use an analytical approximation method. We approximate $g(\theta|\Xi,x_1,\ldots,x_N)$ as the standard large-sample normal approximation. The approximate posterior mean is given by the mode $\tilde{\theta}$ of the posterior distribution. The posterior mode $\tilde{\theta}$ can be obtained by using Newton's method to maximize the logarithm of the joint mass/density function:

$$\tilde{\theta} = \arg\max_\theta\left\{\log\big(p(x_1|\theta,\Xi)\cdots p(x_N|\theta,\Xi)g(\theta|\Xi)\big)\right\} \quad (7)$$

The approximate posterior covariance is equal to the inverse of the Hessian matrix of second derivatives (which can be obtained in closed form):

$$\tilde{\Sigma} = \left[-D_\theta^2 \log\big(p(x_1|\theta,\Xi)\cdots p(x_N|\theta,\Xi)g(\theta|\Xi)\big)\right]^{-1}. \quad (8)$$

The normal approximation is accurate for large sample sizes [26]. The posterior density function is always unimodal, and we have observed that the conditional density for $\theta_\xi$ given the other $\theta$'s is not too asymmetric even when the corresponding marginal $T_\xi$ is small. Nevertheless, the quality of the Laplace approximation is a question worthy of further investigation.

## 4. POSTERIOR PROBABILITIES FOR STRUCTURES

To perform inference with multiple structures, it is natural to assign a prior distribution for $\theta$ that is a *mixture* of distributions of the form (7). That is, a structure probability $\pi_\Xi$ and a prior distribution $g(\theta|\Xi)$ is specified for each of the interaction structures under consideration. The structure-specific parameter distribution $g(\theta|\Xi)$ was described above in Section 3. To obtain the prior distribution for structures, we used discussions with neuroscientists and experience fitting similar models to other data sets. We expected all $\theta_\xi$ for single-element $\xi$ to be nonzero. We expected most other $\theta_\xi$ to be zero [27]. The prior distribution we used assumed that the $\theta_\xi$ were zero or nonzero independently, and each $\theta_\xi$ with $|\xi|>1$ had prior probability of .1 of being nonzero. The results of the analysis were insensitive to the (nonzero) prior probability of including singleton $\xi$.

The posterior distribution of $\theta$ given a sample of observations is also a mixture distribution:

---

[5] The sufficient statistic $\bar{T}$ contains $T_\xi$ only for those clusters in $\Xi$. For notational simplicity the dependence of $T$ on the structure $\Xi$ is suppressed.

376  Laskey and Martignon

$$g(\theta|x_1,...,x_N) = \sum_\Xi \pi_\Xi^* g(\theta|\Xi, x_1,...,x_N) \quad (9)$$

where $\pi_\Xi^*$ is the posterior probability of structure $\Xi$ and $g(\theta|\Xi, x_1,...,x_N)$ is the posterior distribution of $\theta$ given structure $\Xi$ as defined in (4) and (5).

The ratio of posterior probabilities for two structures $\Xi_1$ and $\Xi_2$ is given by:

$$\frac{\pi_{\Xi_1}^*}{\pi_{\Xi_2}^*} = \frac{p(\Xi_1|x_1,...,x_N)}{p(\Xi_2|x_1,...,x_N)} = \frac{p(x_1,...,x_N|\Xi_1)}{p(x_1,...,x_N|\Xi_2)} \frac{\pi_{\Xi_1}}{\pi_{\Xi_2}} \quad (10)$$

Thus, the posterior odds ratio is obtained by multiplying the prior odds ratio by the ratio of marginal probabilities of the observations under each of the two structures. This ratio of marginal probabilities is called the Bayes factor [19]. The marginal probability of the observations under structure $\Xi$ is obtained by integrating $\theta'$ out of the joint conditional density:

$$p(x_1,...,x_N|\Xi)$$
$$= \int p(x_1|\theta',\Xi) \cdots p(x_N|\theta',\Xi) g(\theta'|\Xi) d\theta' \quad . \quad (11)$$

We assumed in Section 3 that the joint mass/density function (and hence the posterior density function for $\theta$) was highly peaked about its maximum and approximately normally distributed. The marginal probability $p(x_1,...,x_N|\Xi)$ is approximated by integrating this normal density:

$$\tilde{p}(x_1,...,x_N|\Xi)$$
$$\approx (2\pi)^{d/2} |\tilde{\Sigma}|^{1/2} p(x_1|\tilde{\theta},\Xi) \cdots p(x_N|\tilde{\theta},\Xi) g(\tilde{\theta}|\Xi) \quad (12)$$

The approximation (12) is known as Laplace's approximation [19]; [28].

The posterior expected value of $\theta$ is a weighted average of the conditional posterior expected values. As noted in Section 3, these expected values are not available in closed form. Using the conditional MAP estimate $\tilde{\theta}_\Xi$ to approximate the conditional posterior expected value, we obtain the point estimate

$$\tilde{\theta} = \sum_\Xi \pi_\Xi^* \tilde{\theta}_\Xi \quad . \quad (13)$$

In (13), the value $\tilde{\theta}_\xi = 0$ is used for structures not containing $\xi$. Also of interest is an estimate of the value of $\theta_\xi$ given that $\theta_\xi \neq 0$, which is obtained by dividing $\theta_\xi$ by the sum of the $\pi_\Xi$ for which $\xi \in \Xi$. Equation (8) can be used to approximate the posterior covariance matrix $\tilde{\Sigma}_\Xi$ for $\theta$ conditional on structure $\Xi$. Again, the unconditional covariance can be approximated by a weighted average:

$$\tilde{\Sigma} = \sum_\Xi \pi_\Xi^* \tilde{\Sigma}_\Xi \quad . \quad (14)$$

(Zero variance is assumed for $\theta_\xi$ when $\xi \in \Xi$.) The variance of $\theta_\xi$ conditional on $\xi \in \Xi$ can be obtained by dividing the variance $\tilde{\sigma}_\xi^2$ estimated from (14) by the sum of the $\pi_\Xi$ for which $\xi \in \Xi$.

Computing the posterior distribution (9) requires computing, for each possible structure $\Xi$, the posterior probability of $\Xi$ and the conditional distribution of $\theta$ given $\Xi$. This is clearly infeasible: for $k$ nodes there are $2^k$ activation clusters, and therefore $2^{2^k}$ possible structures. It is therefore necessary to approximate (9) by sampling a subset of structures. We used a Markov Chain Monte Carlo Model Composition (MC$^3$) algorithm [13] to sample structures. A Markov chain on structures was constructed such that it converges to an equilibrium distribution in which the probability of being at structure $\Xi$ is equal to the Laplace approximation to the structure posterior probability $\pi_\Xi^*$. We used a Metropolis-Hastings sampling scheme, defined as follows. When the process is at structure $\Xi$, a new structure $\Xi'$ is sampled by either adding or deleting a single cluster. The cluster $\xi'$ to add or delete is chosen by a probability distribution $\rho(\xi'|\Xi)$.[6] The proposed addition or deletion is then either accepted or rejected, with the probability of acceptance given by:

$$\min\left\{1, \frac{p(x_1,...,x_N|\Xi')\pi_{\Xi'}\rho(\xi'|\Xi)}{p(x_1,...,x_N|\Xi)\pi_\Xi \rho(\xi'|\Xi')}\right\} \quad (15)$$

## 5. APPLICATION OF THE METHODS

There is growing consensus that processing in the brain is organized in functional groups of neurons. Following Hebb [2], these groups are commonly referred to as "cell assemblies." One operational definition for the cell assembly has been particularly influential: near-simultaneity or some other specific timing relation in the firing of the participating neurons. Such temporal coherence is at least in principle important to brain function: if two neurons converge on a third one, their synaptic influence is much larger for near-coincident firing, due to spatio-temporal summation in the dendrite. Thus, synchrony of firing is directly available to the brain as a potential neural code [4-5], [29-30].

In pursuit of experimental evidence for cell assembly activity in the brain, physiologists thus seek to observe the activities of many separate neurons simultaneously, preferably in awake, behaving animals. These "multi-neuron activities" are then inspected for possible signs of interactions between neurons. Results of such analyses may be used to draw inferences regarding the processes taking place within and between hypothetical cell assemblies. The conventional approach is based on the use of crosscorrelation techniques, usually applied to the activity of pairs (sometimes triplets) of neurons recorded under appropriate stimulus conditions. The result is a time-averaged measure of the temporal correlation among the spiking events of the observed neurons under those conditions. Recently it is becoming possible to examine larger groups of neurons and to study the dynamic properties of the firing correlation between neurons in fine detail. Application of these new measures has revealed interesting instances of time- and context-dependent synchronization dynamics in different cortical areas, particularly in awake, behaving animals. Recent investigations have focused on the detection of individual instances of synchronized activity: "unitary events," consisting of precise spike patterns in multiple-neuron activity, occurring more frequently than expected by chance. The models reported in this paper were developed for the purpose of detecting such unitary events.

---

[6]To improve acceptance probabilities, we modified the distribution $\rho(\xi'|\Xi)$ as sampling progressed. The sampling probabilities were bounded away from zero to ensure convergence.

We applied our models to data from an experiment in which spiking events among groups of neurons were analyzed through multi-unit recordings of 6-16 neurons in the cortex of Rhesus monkeys. The monkeys were trained to localize a source of light and, after a delay, to touch the target from which the light blink was presented. At the beginning of each trial the monkeys touched a "ready-key", then the central ready light was turned on. Three to six seconds later, a visual cue was given in the form of a 200-ms light blink coming from either the left or the right. Then, after a delay of 1 to 32 seconds, the color of the ready light changed from red to orange and the monkeys had to release the ready key and touch the target from which the cue was given.

The spiking events (in the 1 millisecond range) of each neuron were encoded as a sequence of zeros and ones, and the activity of the whole group was described as a sequence of configurations or vectors of these binary states. Since the method presented in this paper does not take into account temporal correlation or nonstationarity, the experimenter provided data corresponding to stationary segments of the trials, which are also those presenting the least temporal correlation, corresponding to intervals of 2000 milliseconds around the ready-signal. He adjoined these 94 segments and formed a data-set of 188,000 milliseconds. The data were then binned in time windows of 40 milliseconds. The choice of window was also chosen in discussion with the experimenter. The frequencies of configurations of zeros and ones in these windows are the data used for analysis in this paper. For computational reasons we selected a subset of six of the eight neurons for which data were recorded.

We analyzed recordings prior to the ready-signal separately from data recorded after the ready-signal. Each of these data sets is assumed to consist of independent trials from a model of the form (2). We ran the $MC^3$ model search algorithm described in Section 4 for 15,000 iterations. Posterior probabilities, estimated $\theta_\xi$, and estimated $\sigma_\xi$ are shown in Tables 1 and 2 for all interactions $\xi$ with estimated posterior probability of at least .1 (i.e., interactions for which the posterior probability is at least as large as the prior probability). Posterior probabilities for interactions were estimated in two different ways: by frequencies collected over the 15,000 iterations of the $MC^3$ algorithm and by normalizing probabilities for the 100 highest-probability structures enumerated during the model search. The results shown in Table 1 and Table 2 indicate that these two estimation methods yield similar estimates. Also shown in Table 1 and Table 2 are estimates for the interaction strength parameters $\theta_\xi$ conditional on $\theta_\xi \neq 0$. The estimate $\bar{\theta}_\xi$ is a weighted average of the MAP estimates (7). Structures included in the weighted average were those from the 100 most probable models that include $\theta_\xi$. Structure weights are proportional to the structure posterior probability.

| Cluster $\xi$ | Posterior Probability of $\xi$ (Frequency) | Posterior Probability of $\xi$ (Best 100 models) | MAP estimate $\bar{\theta}_\xi$ | Standard Deviation of $\theta_\xi$ |
|---|---|---|---|---|
| 1 | 1.00 | 1.00 | -1.52 | 0.06 |
| 2 | 1.00 | 1.00 | -1.73 | 0.07 |
| 3 | 1.00 | 1.00 | -3.13 | 0.15 |
| 4 | 1.00 | 1.00 | -0.82 | 0.06 |
| 5 | 1.00 | 1.00 | -2.76 | 0.10 |
| 6 | 1.00 | 1.00 | -0.83 | 0.06 |
| 4,6 | 0.98 | 1.00 | 0.49 | 0.11 |
| 2,3,4,5 | 0.33 | 0.36 | 2.34 | 0.67 |
| 3,4,6 | 0.22 | 0.15 | 0.80 | 0.27 |
| 2,3,4,5,6 | 0.16 | 0.13 | 2.53 | 0.88 |
| 1,4,5,6 | 0.16 | 0.10 | 6.67 | 1.19 |
| 3,4 | 0.12 | 0.08 | 0.63 | 0.23 |

Table 1: Results for Pre-Ready Signal Data
Effects with Posterior Probability > 0.1

| Cluster $\xi$ | Posterior Probability of $\xi$ (Frequency) | Posterior Probability of $\xi$ (Best 100 models) | MAP estimate $\theta_\xi$ | Standard Deviation of $\theta_\xi$ |
|---|---|---|---|---|
| 1 | 1.00 | 1.00 | -1.03 | 0.06 |
| 2 | 1.00 | 1.00 | -2.54 | 0.10 |
| 3 | 1.00 | 1.00 | -3.86 | 0.24 |
| 4 | 1.00 | 1.00 | -0.40 | 0.05 |
| 5 | 1.00 | 1.00 | -3.06 | 0.12 |
| 6 | 1.00 | 1.00 | -0.50 | 0.05 |
| 3,4 | 0.86 | 0.94 | 1.00 | 0.27 |
| 2,5 | 0.25 | 0.18 | 0.98 | 0.34 |
| 1,4,5,6 | 0.18 | 0.13 | 1.06 | 0.36 |
| 1,4,6 | 0.15 | 0.08 | 0.38 | 0.13 |

**Table 2: Results for Post-Ready Signal Data: Effects with Posterior Probability > 0.1**

The estimate $\tilde{\sigma}_\xi$ is obtained from the sum of "within" and "between" variance components:

$$\tilde{\sigma}_\xi^2 = \sum_{\xi \subseteq \Xi} \tilde{p}_\Xi \left( \tilde{\sigma}_\xi^2 + \left( \tilde{\theta}_\xi - \tilde{\theta}_\xi \right)^2 \right) \quad (16)$$

The sum is over the 100 most probable enumerated structures that include $\xi$; the weight $\tilde{p}_\Xi$ is proportional to the posterior probability of $\Xi$ and normalized to sum to 1; the within-structure variance component is a weighted average of the appropriate diagonal elements of (8); and the between-structure variance component is the sum of squared deviations of the structure-specific parameter estimates about their mean.

The data analysis confirms the prior expectation that not many interactions would be present [27]. There was a high probability second-order interaction in each data set: $\xi_{4,6}$ in the pre-ready data and $\xi_{3,4}$ in the post-ready data. In the pre-ready data, a fourth-order interaction $\xi_{2,3,4,5}$ had posterior probability about 1/3 (this represents approximately three times the prior probability of .1). The tables show a few other interactions with posterior probability larger than their prior probability.

Comparison of Tables 1 and 2 reveals some similarities and some important differences. First, there is strong evidence for different second-order interactions in the two data sets. The pre-ready data shows strong evidence for the {4,6} interaction; there is no evidence for this interaction in the post-ready data. The post-ready data shows strong evidence for the {3,4} interaction; there is no evidence for this interaction in the pre-ready data. (Although the {3,4} interaction term appears in Table 1, its posterior probability is barely greater than its prior probability, indicating that the data show evidence neither for its presence nor for its absence.) Both these second-order interactions are excitatory. Inspection of the coefficient estimates for the single-element effects reveals values that are similar but most likely not identical for the pre and post-ready data. This suggests that the base level of activation of the six neurons is similar in the pre and post-ready situations. (These coefficients are negative because the overall level of activation in these recordings was low.)

We also applied our models to synthetically generated data. We generated data randomly from a model of the form (2), where we specified $\Xi$ and $\theta_\xi$. Table 3 shows the nonzero $\theta_\xi$ in the model we used. The interaction effects included were the four highest posterior probability interactions from the pre-ready data (see Table 1). The values of $\theta_\xi$ were chosen by computing the maximum a posteriori estimate of $\theta$ for the model with these nonzero effects fitted to the pre-ready data. (Note that the coefficients $\theta_{2,3,4,5}$ and $\theta_{2,3,4,5,6}$ are both considerably smaller than the values from Table 1. These two interaction terms are share many neurons and are negatively correlated with each other. When both appear in the model, both are smaller than when either appears alone.) We used the $\theta_\xi$ from Table 3 to compute $p(x)$ from (2), and then generated 2000 random observations from $p(x)$.

Results from the analysis of synthetic data appear in Table 4. Note that the (4,6) and (3,4,6) interactions had very high posterior probability. The former was estimated quite precisely; the latter to within two posterior standard deviations. Neither the (2,3,4,5) nor the (2,3,4,5,6) interactions was detected by the analysis of the synthetic data (their posterior probabilities were .03 and .04, respectively). This suggests that the magnitude of these interactions was not sufficient to be detected in a sample of this size. No interactions that were not present in the synthetic data model were detected by the data analysis.

| Cluster $\xi$ | $\theta_\xi$ |
|---|---|
| 1 | -1.52 |
| 2 | -1.74 |
| 3 | -3.24 |
| 4 | -0.82 |
| 5 | -2.78 |
| 6 | -0.83 |
| 4,6 | 0.45 |
| 3,4,6 | 0.74 |
| 2,3,4,5 | 1.79 |
| 2,3,4,5,6 | 0.61 |

Table 3: Nonzero Parameter Values for Synthetic Data

We implemented the data analysis algorithm in Xlisp-Stat [31] on a PowerMacintosh, using a native Power Macintosh version of Xlisp-Stat. Because an iterative maximization of the posterior distribution was required in every iteration, each iteration took about 2.7 seconds, with the entire 15,000 iterations requiring about 11 hours of computation time. We are currently investigating ways to make the model search process more efficient.

## 6. CONCLUSION

This paper developed a Bayesian approach to the drawing inferences about the structure and parameters of nonhierarchical loglinear models for activity in neurons. The method was applied to a dataset on neural activity in monkeys. Second-order interactions were detected, and the data showed weak evidence for interactions of order higher than 2. Clear differences in the interaction structure were observed for different attentional states of the monkey.

The Bayesian approach has several important advantages, most notably the straightforward interpretation of results, the ability to incorporate structural uncertainty, and the consequent ability to compare many different structures in a statistically justifiable way. The methods reported in this paper are computationally quite intensive. We are investigating the implementation of the methods on a massively parallel computer. This will make it possible to handle data sets consisting of much larger numbers of nodes. We are also planning simulation studies to examine the quality of the Laplace approximation.

### Acknowledgements

The authors extend grateful acknowledgment to Eilon Vaadia, who put at our disposal data obtained at the Department of Physiology, Hadassah Medical School, Hebrew University, Jerusalem, Israel. Ad Aertsen and Sonja Grün are acknowledged for inspiring discussions that clarified why effects of log-linear models provide a good parameterization for neural interactions. Appreciation is extended to Arne Schwartz for assistance in pre-processing of the raw data. The first author was partially supported by a Career Development Fellowship from the Krasnow Institute at George Mason University. The second author was partially supported by a grant for "Habilitation" (Habilitationsstipendium Nr. 1544) of the Deutsche Forschungs Gemeinschaft. The authors gratefully acknowledge helpful comments from anonymous reviewers of earlier versions of this paper.

| Cluster $\xi$ | Posterior Probability of $\xi$ (Frequency) | Posterior Probability of $\xi$ (Best 100 models) | MAP estimate $\tilde{\theta}_\xi$ | Standard Deviation of $\theta_\xi$ |
|---|---|---|---|---|
| 1 | 1.00 | 1.00 | -1.54 | 0.06 |
| 2 | 1.00 | 1.00 | -1.75 | 0.06 |
| 3 | 1.00 | 1.00 | -3.54 | 0.15 |
| 4 | 1.00 | 1.00 | -0.82 | 0.06 |
| 5 | 1.00 | 1.00 | -2.73 | 0.10 |
| 6 | 1.00 | 1.00 | -0.85 | 0.06 |
| 4,6 | 0.99 | 0.99 | 0.46 | 0.10 |
| 3,4,6 | 0.93 | 0.96 | 1.13 | 0.26 |

Table 4: Results for Synthetic Data

380    Laskey and Martignon4. Abeles, M., H. Bergman, E. Margalit, and E. Vaadia. "Spatiotemporal Firing Patterns in the Frontal Cortex of Behaving Monkeys." *Journal of Neurophysiology* 70, no. 4 (1993): pp. 1629-1638.

5. Abeles, M., and G. Gerstein. "Detecting Spatiotemporal Firing Patterns Among Simultaneously Recorded Single Neurons." *Journal of Neurophysiology* 60 (1988): 904-924.

6. Abeles, M., E. Vaadia, Y. Prut, I. Haalman, and H. Slovin. "Dynamics of Neuronal Interactions in the Frontal Cortex of Behaving Monkeys." *Concepts in Neuroscience* 4, no. 2 (1993): pp. 131-158.

7. Hopfield, J. J. "Neural Networks and Physical Systems with Emergent Collective Computational Abilities." *Proceedings of the National Academy of Sciences, USA* 79 (1984): 2554-2558.

8. Hinton, G., J. McClelland, and D. Rumelhart. "Distributed Representations." In *Parallel Distributed Processing*. MIT Press (1986).

9. Hinton, G., and T. Sejnowski. "Learning and Relearning in Boltzmann Machines." In *Parallel Distributed Processing*. MIT Press (1986).

10. Martignon, L., H. Hasseln, S. Grün, A. Aertsen, and G. Palm. "Detecting the Interactions in a Set of Neurons: A Markov Field Approach." *Biological Cybernetics* (1995):

11. Stewart, L. "Hierarchical Bayesian Analysis Using Monte Carlo Integration: Computing Posterior Distributions when there are Many Possible Models." *The Statistician* 36 (1987): pp. 211-219.

12. Madigan, D., and A. Raftery. "Model Selection and Accounting for Model Uncertainty in Graphical Models using Occam's Window." *Journal of the American Statistical Association* 89, no. 428 (1994): 1535-1546.

13. Madigan, D., and J. York. "Bayesian Graphical Models for Discrete Data," Technical Report 239, Department of Statistics, University of Washington, 1993.

14. Buntine, W. L. "Operations for Learning with Graphical Models," Technical report, RIACS & NASA Ames Research Center, Mail Stop 269-2, Moffett Field, CA 94035, USA, 1995.

15. Cooper, G., and E. Herskovits. "A Bayesian Method for the Induction of Probabilistic Networks from Data." *Machine Learning* 9 (1992): pp. 309-347.

16. Ripley, B. D. "Statistical Ideas for Selecting Network Architectures." Neural Networks: Artificial Intelligence and Industrial Applications, Proceedings of the 1995 S.N.N. Symposium, 1995.

17. Cooper, G. F., and S. Rajagopalan. "A Method for Learning the Structure of a Neural Network from Data." NIPS, 1995.

18. Geiger, D., D. Heckerman, and C. Meek. "Asymptotic Model Selection for Directed Networks with Hidden Variables." NIPS, 1995.

19. Kass, Robert E., and Adrian E. Raftery. "Bayes Factors." *Journal of the American Statistical Association* 90, no. 430 (1995): 773-795.

20. Draper, David. "Assessment and Propagation of Model Uncertainty." *Journal of the Royal Statistical Society, Ser. B* 57 (1995): 45-98.

21. Martignon, L., and K. Laskey. "Statistical Inference Methods for Classifying Higher Order Neural Interactions." 1995.

22. Bishop, Y., S. Fienberg, and P. Holland. *Discrete Multivariate Analysis.* Cambridge, MA: MIT Press, 1975.

23. Whittaker, J. *Graphical Models in Applied Multivariate Statistics*. New York: Wiley, 1990.

24. Dawid, A. P., and S. L. Lauritzen. "Hyper Markov Laws in the Statistical Analysis of Decomposable Graphical Models." *Annals of Statistics* 21 (1993): 1272-1317.

25. Darroch, J. N., S. L. Lauritzen, and T. P. Speed. "Markov Fields and Log-Linear Interaction Models for Contingency Tables." *The Annals of Statistics* 8, no. 3 (1980): pp. 522-539.

26. DeGroot, Morris H. *Optimal Statistical Decisions.* New York, NY: McGraw Hill, 1970.

27. Braitenberg, V., and A. Schüz. "Anatomy of the Cortex: Statistics and Geometry." In *Studies in Brain Function. Springer-Verlag* (1994).

28. Tierney, L., and J. B. Kadane. "Accurate Approximations for Posterior Moments and Marginal Densities." *Journal of the American Statistical Association* 81 (1986): pp. 82-86.

29. Grün, S., A. Aertsen, M. Abeles, G. Gerstein, and G. Palm. "Behavior-related Neuron Group Activity in the Cortex." Proceedings of the 17th Annual Meeting of the European Neuroscience Association, Oxford University Press, 1994.

30. Aertsen, A., M. Abeles, G. Gerstein, and G. Palm. "On the Significance of Coincident Firing in Neuron Group Activity." In *Sensory Transduction. Thieme* (1994): 558.

31. Tierney, L. *Lisp-Stat*. New York: Wiley, 1990.